\documentclass{article}

\usepackage{amsmath,amssymb,amsthm,mathrsfs,amsfonts,dsfont}

\usepackage[utf8]{inputenc} 
\usepackage[T1]{fontenc}    
\usepackage{hyperref}       
\usepackage{url}            
\usepackage{booktabs}       
\usepackage{amsfonts}       
\usepackage{nicefrac}       
\usepackage{microtype}      

\usepackage[utf8]{inputenc} 
\usepackage[T1]{fontenc}    
\usepackage{hyperref}       
\usepackage{url}            
\usepackage{booktabs}       
\usepackage{amsfonts}       
\usepackage{nicefrac}       
\usepackage{microtype}      
\usepackage{amsmath, dsfont}
\usepackage{algorithm}
\usepackage[noend]{algpseudocode}
\usepackage{mathtools}

\DeclarePairedDelimiter\floor{\lfloor}{\rfloor}
\usepackage{wrapfig}

\makeatletter
\def\BState{\State\hskip-\ALG@thistlm}
\makeatother\usepackage[pdftex]{graphicx}
\usepackage{xcolor}
\usepackage{soul}

\newtheorem{thm}{Theorem}[section]

\newtheorem{prop}[thm]{Proposition}

\newtheorem{rque}{Remark} [section]

\newcommand{\E}{\mathds{E}}
\newcommand{\V}{\mathrm{Var}}

\title{A Characterization of Mean Squared Error for Estimator with Bagging }

\author{Martin Mihelich\thanks{Walnut Algorithms, \texttt{martin.mihelich@walnut.ai}}, Charles Dognin\thanks{WVerisk AI, Verisk Analytics, \texttt{charles.dognin@verisk.com}}, Yan Shu\thanks{Walnut Algorithms, \texttt{yan.shu@walnut.ai}}, Michael Blot\thanks{Walnut Algorithms,\texttt{michael.blot@walnut.ai}} }

\begin{document}

\maketitle

\begin{abstract}
    Bagging can significantly improve the generalization performance of unstable machine learning algorithms such as trees or neural networks. Though bagging is now widely used in practice and many empirical studies have explored its behavior, we still know little about the theoretical properties of bagged predictions. In this paper, we theoretically investigate how the bagging method can reduce the Mean Squared Error (MSE) when applied on a statistical estimator. First, we prove that for any estimator, increasing the number of bagged estimators $N$ in the average can only reduce the MSE. This intuitive result, observed empirically and discussed in the literature, has not yet been rigorously proved. Second, we focus on the standard estimator of variance called unbiased sample variance and we develop an exact analytical expression of the MSE for this estimator with bagging. 
    This allows us to rigorously discuss the number of iterations $N$ and the batch size $m$ of the bagging method. From this expression, we state that only if the kurtosis of the distribution is greater than $\frac{3}{2}$, the MSE of the variance estimator can be reduced with bagging. This result is important because it demonstrates that for distribution with low kurtosis, bagging can only deteriorate the performance of a statistical prediction. Finally, we propose a novel general-purpose algorithm to estimate with high precision the variance of a sample.
\end{abstract}

\section{Introduction}
    Since the popular paper \cite{breiman1996bagging}, bootstrap aggregating (or bagging) has become prevalent in machine learning applications. This method considers a number $N$ of samples from the dataset, drawn uniformly with replacement (bootstrapping) and averages the different estimations computed on the different sub-sets in order to obtain a better (bagged) estimator.  In this article, we theoretically investigate the ability of bagging to reduce the Mean Squared Error (MSE) of a statistical estimator with an additional focus on the MSE of the variance estimator with bagging.
    
    In a machine learning context, bagging is an ensemble method that is effective at reducing the test error of predictors. Several convincing empirical results highlight this positive effect \cite{webb2004multistrategy,mao1998case}. However, those observations cannot be generalized to every algorithm and dataset since bagging deteriorates the prediction performance in some cases \cite{grandvalet2004bagging}. Unfortunately, it is difficult to understand the reasons explaining why the behavior of bagging differs from one application to another. In fact, only a few articles give theoretical interpretations \cite{buhlmann2002analyzing,friedman2007bagging,buja2016smoothing}. The difficulty to obtain theoretical results is partially due to the huge number of bootstrap possibilities for the estimator with bagging. For instance, in \cite{buja2006observations, CH2003,BS2018}, authors studied properties of bagging-statistic, which considers all possible bootstrap samples in bagging, and provided a theoretical understanding regarding the relationship between the sample size and the batch size. However, a bagging estimator differs from a bagging statistic and as explain in \cite{buja2006observations}, due to the huge number of sampling if we want to use bagging statistic, in practice bagging estimators are used.   These insufficient theoretical guidelines generally lead to arbitrary choices of the fundamental parameters of bagging, such as the batch size $m$ and the number of iterations $N$. 
    
    In this paper, we investigate the theoretical properties of bagging applied to statistical estimators and specifically the impact of bagging on the MSE of those estimators. We provide three core contributions:
    \begin{itemize}
        \item We give a rigorous mathematical demonstration (proof) that the bias of any estimator with bagging is independent from the number of iterations $N$, while the variance linearly decreases in $\frac{1}{N}$. This intuitive behavior, observed in several papers \cite{breiman1996bagging,liu2018reducing}, has not yet been demonstrated. We also provide recommendations for the choice of the number of iterations $N$ and the batch size $m$. We further discuss the implication in a machine learning context. To demonstrate these points, we develop a mathematical framework which enables us, with a symmetric argument, to obtain an exact analytical expression for the bias and variance of any estimator with bagging. This symmetric technique can be used to calculate many other metrics on estimators, we hope that our findings will enable more research in the area. 
        
        \item We use our framework to further study bagging applied to the standard estimator of variance called unbiased sample variance. This estimator is widely used and a lot of work has been done on specific versions, such as the Jackknife variance estimator \cite{efron1981jackknife} or the variance estimator with replacement \cite{cho2009variance}. In this paper, we derive an exact analytical formula for the MSE of the variance estimator with bagging. It allows us to provide a simple criteria based on the kurtosis of the sample distribution, which characterizes whether or not the bagging will have a positive impact on the MSE of the variance estimator. We find that on average, applying bagging to the variance estimator reduces the MSE if and only if the kurtosis of the sample distribution is above $\frac{3}{2}$ and the number of bagging iterations $N$ is large enough. From this theorem, we are able to propose a novel, more accurate variance estimation algorithm. This result is particularly important because it describes explicit configurations where the application of bagging cannot be beneficial. We further discuss how this result fits into common intuitions in machine learning.
        \item 
        Finally, we provide various experiments illustrating and supporting our theoretical results.
    \end{itemize}
      
    The rest of the paper is organized as follows. In Section~\ref{sec:2} we present the mathematical framework along with our derivation of the MSE for any bagged estimators. In Section~\ref{sec:3}, we focus on the particular case of the bagged unbiased sample variance estimator and we provide a new criteria on the variable kurtosis. In Section~\ref{sec:experiments}, we support our theoretical findings with three experiments.

\section{Bagged Estimators: The More The Better}\label{sec:2}
    In this section we rigorously define the notations before presenting the derivation of the different components of the MSE of a bagged statistical estimator, namely the bias and the variance. We subsequently deduce our first contribution: the more iterations $N$ used in the bagging algorithm, the smaller the MSE is, with a linear dependency.
    
    \subsection{Notations, Definitions}
        A dataset, denoted $\ell=(x_i)_{i\in \{1,...n\}}$, is the realization of an independent and identically distributed sample set, noted $L = (X_1,...,X_n)$, of a variable $X \in  \mathcal{X}$. A statistic of the variable $X$ is noted $\theta(X) \in \mathds{R}$. An estimator of $\theta$, computed from $L$, is noted $\hat \theta(L)$.
        
        By considering $U: \{1,..,m\} \mapsto \{1,..,n\}$, a random function, we denote $L_U = (X_{U(1)},...,X_{U(m)})$ a uniform sampling with replacement from $L$ of size $m$. The random variable $U$ is defined on $\mathcal{U} =\{ (u_k)_{k=1\cdots n^m} \}$, the finite set of all $m$ sized sampling with replacement, of cardinal $n^m$. The bagging method considers $N$ such sampling functions taken uniformly from $\mathcal{U}$, noted $B = (U^1, ...,U^N)$. We can now define the bagging estimator:
            \begin{align}
                \Tilde{\theta}(L, B) = \frac{1}{N}\sum_{i=1}^N \hat{\theta}(L_{U^i})  = \frac{1}{N}\sum_{i=1}^N \hat{\theta}(X_{U^i(1)}, ..., X_{U^i(m)}).
                \label{BaggingEstimator}
            \end{align}            
        
        Usually, the size of the sample sets is $n$. Here, we consider the general case where $\hat \theta$ is a function of a set of any size $m$. The number $N$ of iterations is often set to $N \in [|10,100|]$ (where $[|a,b|]$ represents all the natural numbers between $a$ and $b$) without rigorous justification \cite{breiman1996bagging, dietterich2000experimental, lemmens2006bagging}. We discuss this parameter in the following section.
        
        The average with respect to $L$ is noted $\E_L$, and with respect to $B$ is noted $\E_B$. We assume that  $  \forall i \in [|1,n^m|]$, $\E_L(\hat{\theta}(L_{U^i})^2) < \infty$. Since $B$ is defined on a finite set, we define $\E_{(L,B)}$ which is the expected value taken with respect to the pair of random variables $(L, B)$ and we have $\E_{(L,B)}=\E_L(\E_B)=\E_B(E_L)$.
        
        Therefore, there are $(n^m)^N$ possibilities to draw $N$ sampling functions taken uniformly from $\mathcal{U}$ ($(u_1,..,u_N) \in \mathcal{U}^N$) and at $L$ fixed, the average over $B$ takes the form:
        $$ \E_B(\tilde{\theta}(L,B)) = \frac{1}{(n^m)^N}\sum_{(u_1,..,u_N) \in \mathcal{U}^N} \frac{1}{N}\sum_{i=1}^N \hat{\theta}(L_{u_i}).$$

    \subsection{MSE of Estimators with Bagging}
        In this subsection we state the theorem on the dependence of the bagged estimators in terms of the number of iterations $N$. The central idea of the proof is to count all the $(n^m)^N$ possible bagged estimators and to use a symmetric argument\footnote{For more details, please refer to the complete proof in the supplementary material}. The proof of Theorem \ref{thm1} can be found in the supplementary material. We eventually adapt this framework to the case of regression predictors.

        \begin{thm}\label{thm1}
        For a general statistic $\theta$, there exists two positive terms $F$ and $G$, independent of $N$, such that the MSE of the bagged estimator, $\tilde{\theta}$ defined by~\eqref{BaggingEstimator}, satisfies:
        \begin{align*}
            \mathrm{MSE}(\tilde{\theta}) &= \frac{1}{N}F + G
        \end{align*}
        with
        \begin{align*}
             F& = \E_L\big(\V_U(\hat{\theta}(L_U))\big)\\
             G& = \V_L\big(\E_U(\hat{\theta}(L_U))\big) + \big(\E_L(\E_U(\hat{\theta}(L_U)))-\theta\big)^2
        \end{align*}
        
        where $U$ is a random variable uniformly distributed on $\mathcal{U}$.

        More generally, the following equations hold.
        \begin{enumerate}
            \item $\E_{(L,B)}(\tilde{\theta}(L,B)) = \E_L(\E_U(\hat{\theta}(L_U)))$ ,
            \item $\V_{(L,B)}(\tilde{\theta}(L,B)) =  \frac{1}{N}\E_{L}\big(\V_U(\hat{\theta}(L_U)) \big) + \V_L\big( \E_U(\hat{\theta}(L_U)) \big)  $.
            
        \end{enumerate}
        \end{thm}
        
        \begin{rque}
            $\E_L\big(\V_U(\hat{\theta}(L_U))\big)$,  $\V_L\big(\E_U(\hat{\theta}(L_U))\big)$ and $\big(\E_L(\E_U(\hat{\theta}(L_U)))-\theta\big)^2$ are positive and do not depend on $N$.  We deduce that:
            \begin{enumerate}
                \item The higher the $N$, the lower the MSE.
                \item As $N$ goes to $\infty$, we have: 
                $$\lim_{N\rightarrow \infty}\mathrm{MSE}(\tilde{\theta}) = \V_L\big(\E_U(\hat{\theta}(L_U))\big) + \big(\E_L(\E_U(\hat{\theta}(L_U)))-\theta\big)^2.$$
            \end{enumerate}
        \end{rque}

    \subsection{Bagging for Regression}
        \label{regressionSetup}
        In the regression setup, the variable $X$ considered is a pair (input, label), $X = (Y, T)$ from $\mathcal{Y} \times \mathcal{T}$. For any input $y \in \mathcal{Y}$, the statistic considered is $\theta_y(Y,T) = \E(T|Y=y)$. The prediction is noted $\hat{\theta}_y(L)$ and the MSE of this estimator represents the prediction error that needs to be reduced. The bagging version of the estimator is noted $\tilde{\theta}_y(L, B)$.
        

        The MSE of the bagged predictor is 
            $$\E_{y\sim Y}\left(\E_{(L,B)}\left((\tilde{\theta}_y(L,B)-\theta_y)^2\right)\right).$$
        
        According to Theorem~\ref{thm1}, for every $y$, there exists two positive values 
        $$F_y= \E_L\big(\V_U(\hat{\theta}_y(L_U))\big),$$ and $$G_y = \V_L\big(\E_U(\hat{\theta}_y(L_U))\big) + \big(\E_L(\E_U(\hat{\theta}_y(L_U)))-\theta_y\big)^2,$$ independent of $N$, such that:
            $$\E_{(L,B)}\left(\left(\tilde{\theta}_y(L,B)-\theta_y\right)^2\right)=\frac{1}{N} F_y+G_y.$$
        Therefore,
            $$\mathrm{MSE}(\tilde{\theta})=
        \frac{1}{N}\E_{y \sim Y}(F_y)+\E_{y \sim Y}(G_y).$$
        We deduce that on average, increasing the number of iterations $N$ can only reduce the average of the MSE of the bagged estimator in a regression setting.

        In this section, we demonstrated that the MSE of an estimator with bagging is a linear function of $\frac{1}{N}$ ($=\frac{1}{N}F + G$). The multiplicative component of the dependency, $F$, is positive. Thus, increasing the number of iterations $N$ can only improve the accuracy of the estimator. This fact has been observed in several empirical studies \cite{breiman1996bagging,sorokina2007additive}, but to our knowledge has never been rigorously proved. $G$ represents a lower bound on the MSE of the estimator with bagging. This means that a sufficient $N$ ensures that the term $\frac{1}{N}F$ is negligible compared to $G$. Moreover, if $G$ is bigger than the MSE of the estimator without bagging, then bagging only deteriorates the precision of the estimator. This deterioration was observed empirically, but the reasons had not yet been theoretically explained \cite{grandvalet2004bagging,skurichina1998bagging}. This result holds for machine learning algorithms in the regression setup as well. In the following section, we apply this general framework to the specific case of the unbiased sample variance estimator.

\section{Sample Variance Estimator}\label{sec:3}
    In this section, we study the specific case of the bagged unbiased sample variance estimator. Applying the above framework on this particular case, we are able to deduce precise constraints under which using bagging improves on average the MSE of the unbiased sample variance estimator. We derive from this result, given in Theorem~\ref{thm3}, a criteria expressed in terms of the kurtosis $\kappa$ of the sample distribution, the batch size $m$ and the number of iterations $N$. We eventually propose at the end of the section, a novel algorithm which provides a more accurate variance estimation if those criteria are met. Carrying the notation of the precedent section, we assume without loss of generality that $\E(X)=0$ by taking $X:=X-\E(X)$. We denote $\mu_2$ the second moment of the centered random variable and $\mu_4$ the forth moment. Since $X$ is centered, $\mu_2$ is the variance of $X$ and $\mu_4/\mu_2^2$ is the kurtosis of $X$. 
    In the rest of the section $\theta(X) = \mu_2$ is the variance of $X$ and we note $\hat{\theta}(L) = \hat{v}(L) =  \frac{1}{n-1} \sum_{i=1}^n (X_i - \frac{1}{n}\sum_{i=1}^nX_i )^2$. The version of this estimator with bagging follows definition (\ref{BaggingEstimator}) and is noted $\tilde{v}(L, B)$. 

    \subsection{Bias and Variance of the Sample Variance Estimator with Bagging}
        The following theorem gives an exact expression of the bias and variance of the variance estimator with bagging.
        \begin{thm}\label{thm2}
            Let $X$ be a random variable with finite fourth moment and satisfying $\E(X)=0$. 
            The bagged variance estimator $\tilde{v}(L,B)$ with batch size $m$ and number of bagging iteration $N$ satisfies:
            \begin{enumerate}
                \item $\E_{(L,B)}(\tilde{v}(L,B)) = \frac{n-1}{n} \mu_2,$
                \item $\V_{(L,B)}(\tilde{v}(L,B)) = \frac{1}{N}\E_L\big(\V_U(\hat{v}(L_U))\big) + \V_L\big(\E_U(\hat{v}(L_U))\big);$
            \end{enumerate}
            where 
            \begin{align}\label{EB}
                \E_L\big(\V_U(\hat{v}(L_U))\big)& = \frac{n-1}{nm(m-1)}\left(3m - 3 + \frac{n^2-2n+3}{n^2}(6-4m)\right)\mu_2^2 \nonumber\\
                & + \frac{n-1}{nm(m-1)}\left(m-1+\frac{n-1}{n^2}(6-4m)\right)\mu_4,
            \end{align}
            and 
                $$\V_L\big(\E_U(\hat{v}(L_U))\big) = \frac{(3-n)(n-1)}{n^3}\mu_2^2 + \frac{(n-1)^2}{n^3}\mu_4.$$
            Moreover, the MSE of the variance estimator with bagging is:
            $$ \mathrm{MSE}(\tilde{v}(L,B)) = \V_{(L,B)}(\tilde{v}(L,B)) + \frac{1}{n^2}\mu_2^2.$$
        \end{thm}
        The proof of Theorem~\ref{thm2} can be found in the supplementary material.
        

    \subsection{Asymptotic Analysis and Comparison of Estimators}
        In this section, we analyze the asymptotic behavior of the bagged estimator with respect to the parameters $m$ and $N$. We also compare the bagged estimator and the non-bagged estimator. Recall that for the standard variance estimator, the MSE is known(see \cite{RS02}). 
        \begin{prop}\label{propvL}
        Assuming that $X$ has a finite fourth moment, the variance of the standard sample variance estimator $\hat{v}(L)$ is given by:
            $$\V_L(\hat{v}(L)) = \frac{3-n}{n(n-1)}\mu_2^2+\frac{1}{n}\mu_4.$$
        Since this estimator is unbiased, it holds:
            \begin{align*}
                \mathrm{MSE}(\hat{v}(L)) &= \V_L(\hat{v}(L))\\
                &= \frac{3-n}{n(n-1)}\mu_2^2+\frac{1}{n}\mu_4.
            \end{align*}
        \end{prop}
        
        Proposition (\ref{propvL}) combined with Theorem~\ref{thm2} enables us to compare the MSE of $\hat v$ and $\tilde{v}(L,B)$ and we find that:
        \begin{align*}
            &\mathrm{MSE}(\tilde{v}(L,B))-\mathrm{MSE}(\hat v)\nonumber \\
        &=\frac{1}{Nm}\left(\mu_4-\mu_2^2\right)+\frac{1}{n^2}\left(-2\mu_4+3\mu_2^2\right)+o\left(\frac{1}{Nm}+\frac{1}{n^2}\right).
        \end{align*}
        
        We state the following theorem.
        
        \begin{thm}\label{thm3}
             As $n$ tends to $+ \infty$, bagging reduces on average the MSE of the variance estimator if and only if:
            \begin{equation}\label{eqfinal}
            -2\mu_4+3\mu_2^2<0,
            \end{equation}
            and 
            \begin{equation}\label{eqfinalN}
                N>\frac{\mu_4-\mu_2^2}{2\mu_4-3\mu_2^2} \frac{n^2}{m}.
            \end{equation}
        \end{thm}
        Therefore, bagging should be used when \eqref{eqfinal} and \eqref{eqfinalN} are satisfied. Thus $N$ and $m$ should be carefully chosen. Moreover, the gain obtain by using bagging is in $O(\frac{1}{n^2})$.  The equation \eqref{eqfinal} can be rewritten as $\kappa > \frac{3}{2}$, with $\kappa$ the kurtosis of $X$. 
        
        Remark that if $m\leq n$, then 
         \begin{equation*}
            \frac{\mu_4-\mu_2^2}{2\mu_4-3\mu_2^2} \frac{n}{m}> \frac{1}{2}.
        \end{equation*}
        We deduce that the number of iterations $N$ should be at least $n/2$ in this case. As previously mentioned, the number of iterations $N$ of bagging is often chosen as $N \in [|10,100|]$ without theoretical justifications \cite{fazelpour2016investigating}. Theorem \ref{thm3} gives, for the sample variance estimator, a minimum number of iterations $N$ above which bagging improves the estimation.
        
        Note that the condition on the kurtosis ($\kappa > \frac{3}{2}$) is not restrictive. In fact, many classical continuous distributions satisfy this condition. For example, $\kappa=3$ for a normal distribution, $\kappa=4.2$ for a logistic distribution, $\kappa=1.8$ for a uniform distribution, $\kappa=9$ for an exponential distribution. We now propose a novel algorithm to estimate the variance of a sample.

    \subsection{Algorithm for Higher Precision Variance Estimation}
        We deduce from the preceding theoretical results a simple algorithm to estimate the variance of a sample with higher precision. To simplify the procedure, we choose to take the batch size $m$ equal to the full sample size $n$. Let $q$, an integer greater or equal to 1, denote a parameter used to control the quality of the resulting estimation. The higher  the $q$, the higher $N$; and as a result of the Theorem~\ref{thm1} and Theorem~\ref{thm3}, the better the estimation is (at the cost of an increasing computation time). We suggest taking $q \approx 2-10$. $\floor{.}$ is the floor function. The reader can find an analysis of the algorithmic complexity in the supplementary material.
        
        \label{algo}
        \begin{algorithm}
        \caption{Variance Estimation Algorithm}\label{euclid}
        \begin{algorithmic}[1]
        \Procedure{VarianceEstimator}{$L=(x_1, \cdots, x_n), n, q$}
        \State $\bar{x} \gets \frac{1}{n}\sum_{i=1}^{n}x_i$
        \State $\hat{\mu}_4 \gets \frac{1}{n}\sum_{i=1}^{n}({x_i} - \bar{x})^4$
        \State $\hat{v} \gets \frac{1}{n-1}\sum_{i=1}^{n}(x_i-\bar{x})^2$
        \If{$-2\hat{\mu}_4+3\hat{v}^2<0$}
        \State $N \gets q \times (\floor{\frac{\hat{\mu}_4-\hat{v}^2 }{2\hat{\mu}_4-3\hat{v}^2 }n} + 1)$
        \For{$i \in \{1, \cdots, N\}$}
        \State Draw with replacement $n$ points from the dataset
        \State $\hat{v}_{i} \gets \hat{v}(x_{U(1)}, \cdots, x_{U(n)})$ 
        \EndFor
        \State $\tilde{v} \gets \frac{1}{N}\sum_{i=1}^N\hat{v}_{i}$ 
        \State $\hat{v} \gets \tilde{v}$
        \EndIf
        \State \textbf{return} $\hat{v}$ 
        \EndProcedure
        \end{algorithmic}
        \end{algorithm}

\section{Experiments}\label{sec:experiments}

    In this section we present various experiments illustrating and supporting the previous theoretical results. The section begins with experiments on general estimator, with a focus on regression estimators as presented in section~\ref{regressionSetup}. Then, we present empirical experiments about the bagging version of the sample variance estimator, supporting our theoretical results of section~\ref{sec:3}. For the experiments, we used Python and specifically the Scikit-Learn \cite{pedregosa2011scikit}, Numpy \cite{van2011numpy}, Scipy \cite{scipy} and Numba \cite{lam2015numba} libraries\footnote{Our code will be released on Github for reproducibility.}. More details on the experimental setup can be found in the supplementary material.  
    
    \subsection{Variance of Bagged Estimators in Regression}
        In order to empirically validate Theorem~\ref{thm1} we performed experiments on two regression predictors: the linear regression and the decision tree regression. In each case, we measured the MSE on the test set after training each model on bagged samples, varying the parameter $N$ of bagging iterations.
        We generated toy regression datasets using the Scikit-Learn library. Each dataset, has a specific amount of Gaussian noise ($\sigma$) applied to the output (0.5 or 5). The number of samples (1000) and the size of the feature space dimension (2) remained constant. To obtain the bagged and non-bagged predictors, we trained the regressors on 5\% of the data and tested it on the remaining 95\%. This partition was chosen to enable a fast training of the growing number of bagged predictors. We are interested in showing empirically that our theoretical relation holds, not to achieve good absolute results. Using the notations from \ref{regressionSetup}: the average $\E_{y}$ was taken over the test set of size $950$. $\E_{B}$ was taken over $100$ trials and $\E_{L}$ was taken over 10 trials. As expected, we observe that the MSE of those estimators decreases at a rate of $\frac{1}{N}$. We fit a non-linear function on the resulting datapoints of the form $\hat{y}=\hat{a} + \hat{\frac{b}{N}}$ to better observe this relationship. More details on the experiment setup and hyperparameters used can be found in the supplementary material.

        \vspace{-10pt}
        \begin{figure}[h] 
        \noindent\makebox[\textwidth][c]{
        \begin{minipage}{0.4\linewidth}      \centering
        \includegraphics[width=1\linewidth]{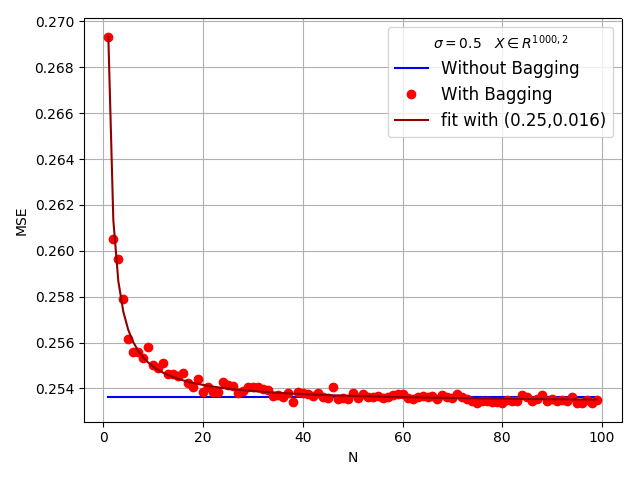} 
        \vspace{1ex}
        \end{minipage}
        \begin{minipage}{0.4\linewidth}
        \centering
        \includegraphics[width=1\linewidth]{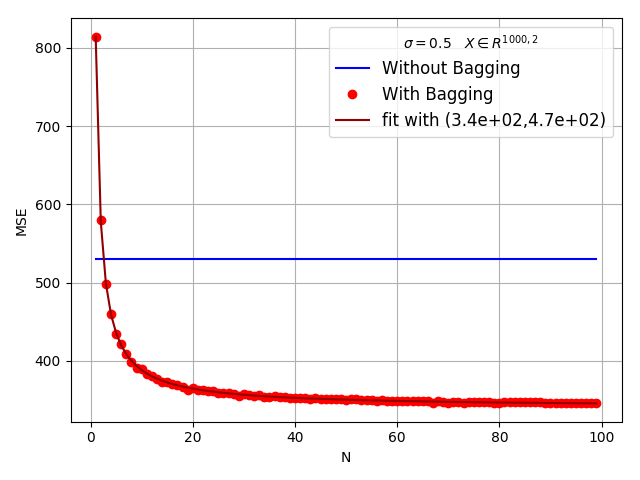} 
        \vspace{1ex}
        \end{minipage} }
        \noindent\makebox[\textwidth][c]{
        \begin{minipage}{0.4\linewidth}
        \centering
        \includegraphics[width=1\linewidth]{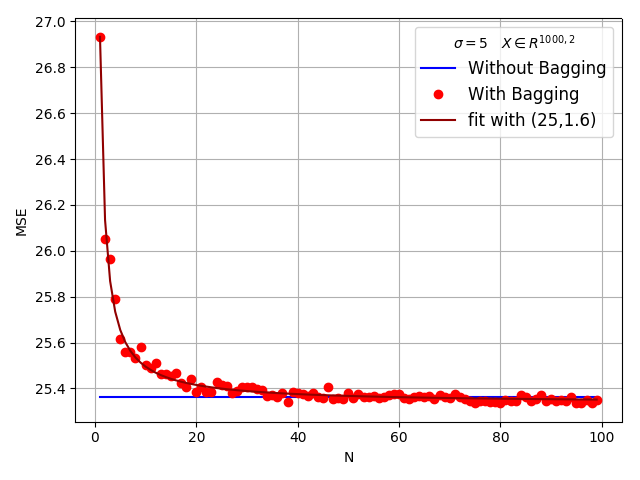} 
        \vspace{1ex}
        \end{minipage}
        \begin{minipage}{0.4\linewidth}
        \centering
        \includegraphics[width=1\linewidth, ]{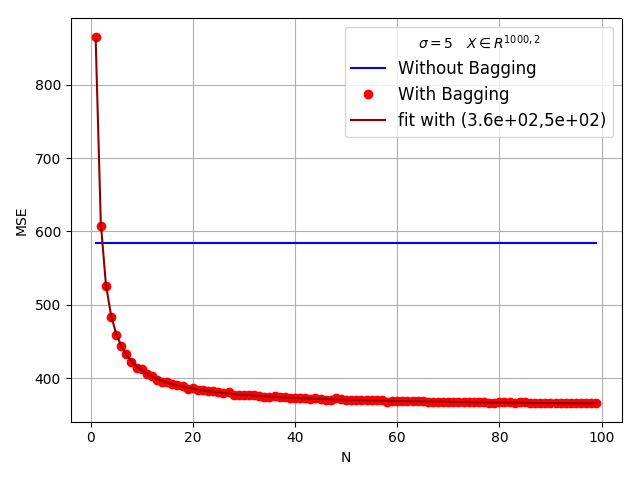} 
        \vspace{1ex}
        \end{minipage} }
        \caption{Linear Regression (Top-Left and Bottom-Left) and Regression Tree (Top-Right and Bottom-Right) Predictors. We observe the $O(\frac{1}{N})$ convergence rate that we demonstrated analytically. We also observe that for the Linear Regression, considered as a stable predictor, bagging does not help. On the other hand, for the Decision Tree Regression, considered as an unstable predictor (following the definition of \cite{breiman1996bagging}, predictors are unstable if a small change in the training set can result in large changes in predictions), bagging helps.}
        \label{fig:fig1}
        \end{figure}

    \subsection{MSE of Bagged and Non-Bagged Unbiased Variance Estimator}
        In this experiment, we test the empirical validity of Theorem~\ref{thm3}. Our condition on the kurtosis comes from the following equality that we proved in the supplementary material: $\mathrm{MSE}(\tilde{v}(L,B))-\mathrm{MSE}(\hat v)\nonumber=\frac{1}{n^2}\left(-2\mu_4+3\mu_2^2\right)+O\left(\frac{1}{Nm}+\frac{1}{n^2}\right)$. Here we set $m=n$. We measured the distance between the MSE of a bagged and non-bagged unbiased sample variance estimator for three classical probability distributions: a standard Gaussian distribution, a uniform distribution on $[-1, 1]$ and a Rademacher distribution. We fixed the number of iterations $N$ to $50$, only varying the sample size $n$, and averaged the estimated variance over $10000$ trials. As expected, it follows the same shape as our condition $(-2\mu_4+3\mu_2^2)$ divided by $n^2$. 
            \vspace{-10pt}
            \begin{figure}[h!] 
              \label{Regression Predictors} 
              \noindent\makebox[\textwidth][c]{
              \begin{minipage}[b]{0.4\linewidth}
                \includegraphics[width=1\linewidth]{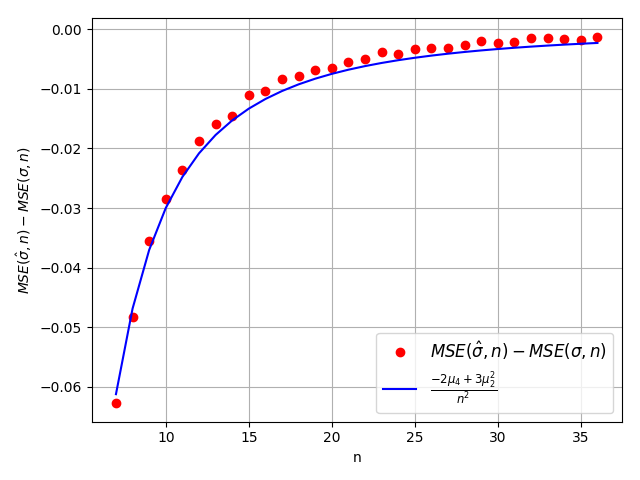} 
                Distance Between MSE of a bagged and non-bagged estimator for a standard Gaussian distribution, averaged over 10000. Here $-2\mu_4+3\mu_2^2=-3.$
                \vspace{4ex}
              \end{minipage}
              \hspace{4ex} 
              \begin{minipage}[b]{0.4\linewidth}
                \includegraphics[width=1\linewidth]{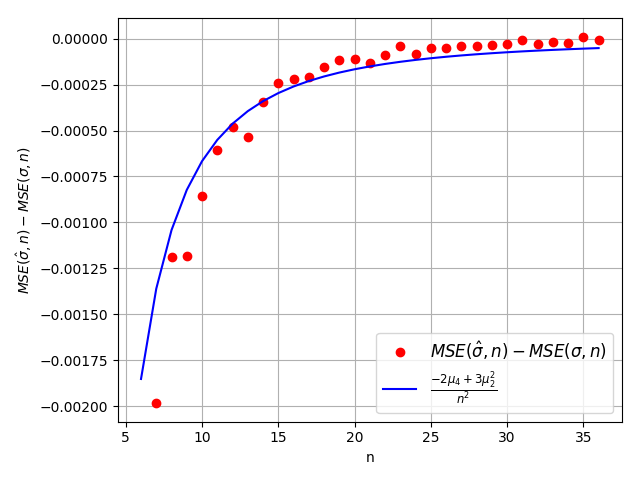} 
                Distance Between MSE of a bagged and non-bagged estimator for a Uniform disribution between -1 and 1, averaged over 10000. Here $-2\mu_4+3\mu_2^2=-\frac{1}{15}.$
                \vspace{4ex}
              \end{minipage}}
                \noindent\makebox[\textwidth][c]{
              \begin{minipage}[b]{0.4\linewidth}
                \includegraphics[width=1\linewidth]{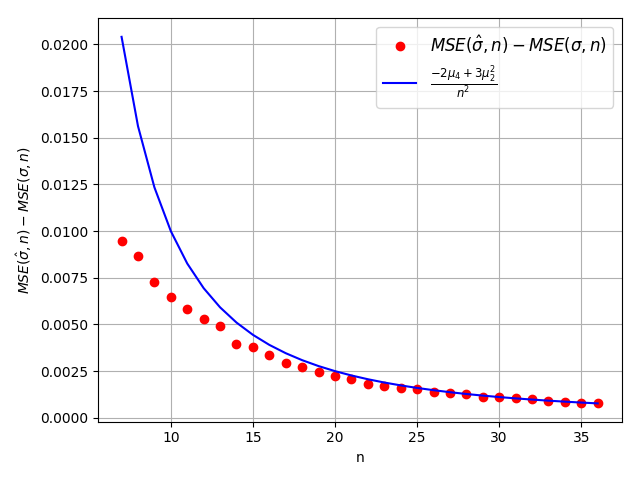} 
                Distance Between MSE of a bagged and non-bagged estimator for a Rademacher disribution, averaged over 10000. Here $-2\mu_4+3\mu_2^2=1.$
                \vspace{4ex}
              \end{minipage}}
            \vspace{-10pt}
             \caption{Usual Distribution Experiments}
            \label{fig:fig2}
        \end{figure}

    \subsubsection{Kurtosis Condition}
\begin{figure}[h!] 
                \centering
                \includegraphics[width=0.5\linewidth]{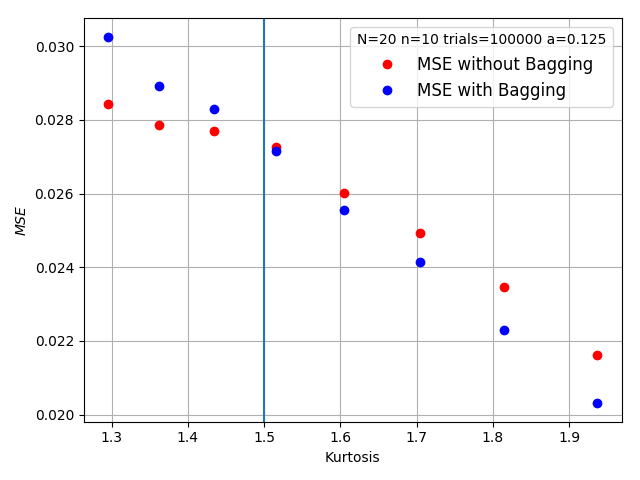} 
             \caption{The MSE of the variance estimator of a distribution with or without bagging with respect to the kurtosis. As expected, we observe that the MSE of the bagged variance estimator becomes lower than the MSE of the non-bagged variance estimator after $\frac{3}{2}$}
            \label{fig:fig3}
            \end{figure}

    In this experiment, we tested the kurtosis condition $\kappa > \frac{3}{2}$. Distributions with kurtosis lower than $\frac{3}{2}$ are unusual as previously mentioned. We designed a distribution whose kurtosis can vary above and below $\frac{3}{2}$ when the parameter $p$ is changing. Let $X$ be a random variable following this distribution. With $p\in [0,1]$, $a>0$ and $p+q=1$: 
        $$P(X=1)=P(X=-1)=\frac{p}{2}.$$ and  
        $$P(X=\sqrt{a})=P(X=-\sqrt{a})=\frac{q}{2}.$$ 
        The kurtosis of this distribution is thus: 
        $$\kappa=\frac{p + qa^2}{(p+qa)^2}.$$ 
        Varying the parameter $p$ between 0 and 1, the kurtosis varies from $1$ to $\infty$ and Figure  \ref{fig:fig3} shows the MSE with and without bagging accordingly. We fixed $N$ to $20$, $n$ to $10$, $a$ to $\frac{1}{8}$, and we averaged over $100000$ trials. As expected, The MSE of the estimator with bagging becomes better than the MSE of the estimator without bagging when the $\frac{3}{2}$ threshold is passed. 

\section{Conclusion}
    In this paper, we theoretically investigate the MSE of estimators with bagging. We prove the existence of a linear dependency of the MSE on $\frac{1}{N}$, $N$ being the number of bagged averaged estimators. This dependency enables us to provide guidelines for setting the parameter $N$. We also explain why bagging is detrimental in some cases. We use the mathematical framework developed Section~\ref{sec:2} to describe the MSE for the sample variance estimator with bagging. It appears that the bagging can reduce the MSE of this estimator if and only if the kurtosis of the distribution is above $\frac{3}{2}$. This condition holds for a large number of classical probability distributions. Using this condition, we eventually propose a novel algorithm for more accurate variance estimation.  
    

\clearpage
\bibliographystyle{unsrt}

\end{document}


\maketitle
\section{Proof of Theorem 2.1}
        
        
        
         Recall that the estimator with bagging is given by:
            $$\tilde{\theta}(L,B) =\frac{\sum_{k=1}^N \hat{\theta}(L_{U^k})}{N},$$
        where $B = (U^1,..,U^N)$ is a set of $N$ sampling with replacement of the dataset $L = (X_1,...,X_n)$, taken independently and uniformly from $\mathcal{U} = \{1,..,n\}^{\{1,..,m\}}  = (u_k)_{k=1..n^m}$. Since we assumed that  $ \forall i \in [|1,n^m|]$, $\E_L(\hat{\theta}(L_{u^i})^2) < \infty$ and $B$ living in a finite set, $\tilde{\theta}$ is well defined and the Fubini's theorem is always available for interchange between different expectations.

        \begin{lemma}
        Let $U$ be a random variable uniformly distributed over $\mathcal{U}$. Then the expected value of the bagged estimator $\tilde{\theta}$ does not depend on $N$ and is equal to $\E_L\E_U(\hat{\theta}(L_U))$. 
        \label{lemma1}
        \end{lemma}
        \begin{proof}
      For $N$ independent identically distributed bagged samples represented by $U^1,...U^N \in \{1,..,n\}^m$, it follows by linearity of the expectation operator and Fubini's theorem that: 
        \begin{align*}
            \E_{(L,B)}(\tilde{\theta}(L, B)) &= \E_L\left(\E_B\left(\frac{1}{N}\sum_{i=1}^N\hat{\theta}(L_{U^i})\right) \right) \\
            &= \E_L\left(\frac{1}{N}\sum_{i=1}^N\E_{U^i}\left(\hat{\theta}(L_{U^i})\right) \right)\\
            &= \E_L\left(\frac{1}{N}\sum_{i=1}^N\E_{U}\left(\hat{\theta}(L_{U})\right) \right)\\
            &= \E_L(\E_U(\hat{\theta}(L_{U}))).    
        \end{align*}
        \end{proof}
        
        Now we turn to the variance. Previous lemma \ref{lemma1} computes the expectations, therefore we only need to compute the second moment. 
        {\begin{lemma}
        The expected value of the bagged estimator satisfies: 
        \begin{equation*}
        \E_{(L,B)}(\tilde{\theta}^2) = \E_L(\E_U(\hat{\theta}(L_U))^2) + \frac{1}{N}\E_L(\V_U(\hat{\theta}(L_U))),
        \end{equation*}
        where $U$ is a random variable uniformly distributed over $\U$ and 
        $$\V_U(\hat{\theta}(L_U))) = \E_U(\hat{\theta}(L_U)^2) - \E_U(\hat{\theta}(L_U))^2.$$
        \label{lemma2}
        \end{lemma}}
        
        \begin{proof}
        {Remark that:
        $$\E_{(L,B)}(\tilde{\theta}^2)= \E_L[\E_B(\tilde{\theta}^2)],$$
        and 
        $$
        \E_B(\tilde{\theta}^2) = \E_B\left(\frac{1}{N}\sum_{i=1}^N\hat{\theta}(L_{U^i})\right)^2.
        $$
        Since $U^i$ is uniformly distributed on $\U$, $\E_B(\tilde{\theta}^2)$ is a symmetric polynomial with respect to $\{\hat{\theta}(L_{u_i})\}_{u_i\in\U}$. Thus there exists two constants $C_1$ and $C_2$ such that for any function $\hat{\theta}$,
        \begin{equation}
        \E_B({\tilde{\theta}}^2)= C_1\left(\sum_{i=1}^{n^m} \hat{\theta}(L_{u_i})^2\right)+C_2\left(\sum_{(i,j)\ i\neq j} \hat{\theta}(L_{u_i}) \hat{\theta}(L_{u_j})\right).
        \label{eqegal}
        \end{equation}
        Let $\hat{\theta} =\hat{\theta}_1 \equiv 1$, it holds 
        \begin{equation}\label{eqegal1}
            1 = C_1 n^m + C_2 n^m(n^m-1).
        \end{equation}}
        
        {Now define $\hat{\theta} =\hat{\theta}_2$, such that $\hat{\theta}_2(L_{u_1}) = 1$ and $\hat{\theta}_2(L_{u_i}) = 0$ for all $i\neq 1$. The right hand side of equation \eqref{eqegal} equals to $C_1$ and we need to compute the value of left hand side. 
        Observe that 
        $$\tilde{\theta}_2 = \frac{1}{N}\sum_{i=1}^N \hat{\theta}_2(L_{U^i}),$$
        where $U^i$ is uniformly and independently distribute on $\mathcal{U}$. Thus
        $\{\hat{\theta}_2(L_{U^i})\}_i$ is a family of independent Bernoulli with parameter $1/n^m$. The second moment of $N \tilde{\theta}_2$ is thus binomial distributed with parameter $(N, 1/n^m)$. Recall that the second moment of a binomial distribution with parameter $(N,p)$ equals to $N(N-1)p^2+Np$. Therefore:
        $$ \E_B(\tilde{\theta}_{2}^2) = \frac{1}{N^2}\left(\E_B(N \tilde{\theta}_{2})^2\right) =  \frac{1}{N^2} \left(N(N-1)n^{-2m}+Nn^{-m}\right).$$
        Simplifying equation \eqref{eqegal} we have:
        $$C_1 = \frac{N-1}{N}n^{-2m} + \frac{1}{N}n^{-m}.$$
        Together with \eqref{eqegal1}, we deduce :
        $$C_2 = \frac{N-1}{N}n^{-2m}.$$
        Hence, 
        \begin{align*}
        \E_B(\tilde{\theta}^2) & = \frac{N-1}{N}n^{-2m}\left(\sum_i \hat{\theta}(L_{u_i})^2 + \sum_{i\neq j} \hat{\theta}(L_{u_i})\hat{\theta}(L_{u_j})\right) + \frac{1}{N}n^{-m}\sum_i \hat{\theta}(L_{u_i})^2\\
        &= \frac{N-1}{N}\E_U(\hat{\theta}(L_U))^2 + \frac{1}{N}\E_U(\hat{\theta}(L_U)^2)\\
        &=\E_U(\hat{\theta}(L_U))^2 + \frac{1}{N}\V_U(\hat{\theta}(L_U)).
        \end{align*}}
        \end{proof}
        
        Now we are in position to prove theorem:
        \begin{proof}[Proof of theorem]:
        According to lemma \ref{lemma1} and \ref{lemma2}, the variance of the bagging estimator is:
        \begin{align*}
            \V_{(L,B)}(\tilde{\theta}) &= \E_{(L,B)}(\tilde{\theta}^2)-\E_{(L,B)}(\tilde{\theta})^2\\
            &= \E_L\left(\E_U(\hat{\theta}(L_U))^2\right) + \frac{1}{N}\E_L(\V_U(\hat{\theta}(L_U))) -\E_L(\E_U(\hat{\theta}(L_U))^2 \\
            &= \frac{1}{N}\E_L(\V_U(\hat{\theta}(L_U))) + \V_L(\E_U(\hat{\theta}(L_U))).
        \end{align*}
        Hence, the MSE of $\tilde{\theta}$ is : 
        \begin{align*}
            & \mathrm{MSE}(\tilde{\theta}) = \V(\tilde{\theta}) + \mathrm{Bias}(\tilde{\theta}) \\
        &=\frac{1}{N}\E_L\big(\V_U(\hat{\theta}(L_U))\big) + \V_L\big(\E_U(\hat{\theta}(L_U))\big) + \big(\E_L(\E_U(\hat{\theta}(L_U)))-\theta\big)^2.
        \end{align*}
        \end{proof}
        

    \section{Proof of Theorem 3.1}
        We recall that $\hat{v}(L)$ can be rewritten as:
        \begin{equation}\label{eqvar}
          \hat{v}(L) = \frac{1}{n(n-1)}\sum_{i\neq j} (X_i-X_j)^2.
        \end{equation}
        {We begin with some lemmas.
        \begin{lemma}\label{lemmaPQR}
        Let $X$ be a random variable with finite forth moment satisfying $\E(X)=0$. Let $X_1,...,X_n$ be independent copies of $X$ and $L = (X_1,...,X_n)$.  Denote $i,j,k,l$ four distinct index of $\{1,...,n\}$. Then the following statements hold.
        \begin{enumerate}
        \item $\E\left((X_i-X_j)^4\right)=6\mu_2^2+2\mu_4$,
        \item $\E\left((X_i-X_j)^2(X_i-X_k)^2\right)=3\mu_2^2+\mu_4$,
        \item $\E\left((X_i-X_j)^2(X_k-X_l)^2\right)=4\mu_2^2$,
        \end{enumerate}
        where $\mu_4 := \E(X^4)$ is the forth moment of $X$.\\
        Now denote $P = \sum_{i,j} (X_i-X_j)^4$, $Q = \sum_{i,j,k}(X_i-X_j)^2(X_i-X_k)^2$ and $R = \sum_{i,j,k,l}(X_i-X_j)^2(X_k-X_l)^2$. Then the following equations hold:
        \begin{enumerate}
            \item[4.] $\E_L(P) = 3n(n-1)\mu_2^2+n(n-1)\mu_4$,
            \item[5.] $\E_L(Q) = \frac{3}{2}n(n-1)(n-2)\mu_2^2+\frac{1}{2}n(n-1)(n-2)\mu_4$,
            \item[6.] $\E_L(R) = \frac{1}{2}n(n-1)(n-2)(n-3)\mu_2^2$.
        \end{enumerate}
        \end{lemma}}
        \begin{proof}
        The proof of the three first items is immediate by independence of $X_i,X_j,X_k,X_l$.
        
        Proof of item 4. $P = \sum_{i,j} (X_i-X_j)^4$ contains $n(n-1)/2$ terms. Combining the item 1, the equation holds.
        
        Item 5 and item 6 can be deduce by similar arguments.
        \end{proof}
        
        {According to theorem 2.1 , we need to compute $\E_L(\E_U(\hat{v}(L_U)))$, $\E_L\big(\V_U(\hat{v}(L_U))\big)$ and $ \V_L\big(\E_U(\hat{v}(L_U))\big)$. We begin with $\E_U(\hat{v}(L_U))$ and $\E_U(\hat{v}^2(L_U))$.}
        
        {\begin{lemma}\label{lemmakeyquantities}
        Let $X$ be a random variable with finite forth moment satisfying $\E(X)=0$. Let $X_1,...,X_n$ be independent copies of $X$ and $L = (X_1,...,X_n)$. Let $U$ be a random variable uniformly distributed over $\mathcal{U}$. Then the following equations hold:
        \begin{align}
        \E_L(\E_U(\hat{v}(L_U))) &= \frac{n-1}{n}\mu_2,\\
        \E_L(\E_U(\hat{v}(L_U))^2)& = \frac{(n-1)(n^2-2n+3)}{n^3}\mu_2^2 + \frac{(n-1)^2}{n^3}\mu_4,\\
        \E_L(\E_U(\hat{v}(L_U)^2)) &= \frac{n-1}{nm(m-1)}\left(3m - 3 + \frac{n^2-2n+3}{n^2}(m-2)(m-3)\right)\mu_2^2 \nonumber\\
        &+ \frac{n-1}{nm(m-1)}\left(m-1+\frac{n-1}{n^2}(m-2)(m-3)\right)\mu_4. 
        \end{align}
        \end{lemma}}
        
        {\begin{proof}
        Item 1. Since $U$ is uniformly distributed on $\mathcal{U}$, it holds
        \begin{align}\label{eqEBvf}
            &\E_U(\hat{v}(L_U)) = \frac{1}{n^m}\sum_{i = 1}^{n^m}\hat{v}(L_{u_i}) \nonumber \\
            &=\frac{1}{n^m}\sum_{i = 1}^{n^m} \frac{1}{m(m-1)}\sum_{j\neq k} (X_{u_i(j)}-X_{u_i(k)})^2.
        \end{align}
        Remark that $\sum_{i = 1}^{n^m}\sum_{j\neq k, j,k\leq m} (X_{L_{u_i}(j)}-X_{L_{u_i}(k)})^2$ is a symmetric polynomial of $(X_j-X_k)^2$, there exists a constant $A$ such that
        $$\sum_{i = 1}^{n^m}\sum_{j\neq k, j,k\leq m} (X_{u_i(j)}-X_{u_i(k)})^2 = A\sum_{j\neq k, j,k\leq n} (X_j-X_k)^2.$$
        $A$ is thus the coefficient of $(X_1-X_2)^2$ in both side.
        For $U\in \U$, denote $n_1(U)$ and $n_2(U)$ the number of $X_1$ and the number of $X_2$ in the sample of bagging $\{X_{U(j)}\}_{j=1,...,m}$. Then the coefficient of $(X_1-X_2)^2$ in $\sum_{j\neq k, j,k\leq m} (X_{U(j)}-X_{U(k)})^2$ is $n_1(U)n_2(U)$. Therefore,
        \begin{equation}\label{eqA}
            A = \sum_{j = 1}^{n^m} n_1(u_j)n_2(u_j).
        \end{equation}
        Consider the function 
        $$G(X_1,...,X_n) = \E_U\left(\exp\left(\sum_{i=1}^m X_{U(i)}\right)\right). $$
        Since $G$ can also be rewritten as
        $$G = \frac{1}{n^m}\sum_{j = 1}^{n^m} \prod_{i=1}^n \exp(n_i(u_j)X_i), $$
        we deduce that 
        $$\sum_{i = 1}^{n^m} n_1(u_i)n_2(u_i) = n^m \partial_1\partial_2 G(0,...,0).$$
        On the other hand, uniformly pick a $U$ out of $\U$ is equivalent to uniformly pick the value of $U(1), U(2),\cdots, U(m)$ out of $\{1,\cdots,n\}$ independently. Hence, 
        \begin{align*}
            & G(X_1,...,X_n) = \E_U\left(\exp\left(\sum_{i=1}^m X_{U(i)}\right)\right) \\
            & = \prod_{i=1}^m \E_U\left(\exp\left(X_{U(i)}\right)\right) = \E_B\left(\exp\left(X_{u}\right)\right)^m,
        \end{align*}
        where $u$ is uniformly distributed on $\{1,\cdots,n\}$. Therefore 
        \begin{equation}\label{fctchara}
        G(X_1,...,X_n)= \E_U\left(\exp\left(X_u\right)\right)^m = \frac{1}{n^m}\left(\sum_{i = 1}^n exp(X_i)\right)^m,
        \end{equation}
        and 
        $$n^m \partial_1\partial_2 G(0,...,0) = m(m-1)n^{m-2}.$$
        Together with equation \eqref{eqA}, it holds:
        $$A = m(m-1)n^{m-2}.$$
        Plugin into equation \eqref{eqEBvf}, we have
        \begin{equation}\label{eqEBvffinal}
            \E_U(\hat{v}(L_U)) = \frac{1}{n^2} \sum_{j\neq k} (X_{j}-X_{k})^2.
        \end{equation}
        Recall that the standard variance estimator is an unbias estimator and can be written as:
        $$\hat{v}(L)) = \frac{1}{n(n-1)} \sum_{j\neq k} (X_{j}-X_{k})^2,$$
        we deduce that 
        $$\E_L(\E_U(\hat{v}(L_U))) = \frac{n-1}{n}\mu_2.$$
        Item 2. By equation \eqref{eqEBvffinal} and a simple development,
        \begin{align}\label{eqvL0}
            \E_U(\hat{v}(L_U))^2 & = \frac{1}{n^4}\left( \sum_{j\neq k} (X_{j}-X_{k})^2\right)^2 \nonumber\\
            & =\frac{1}{n^4} (P + 2Q + 2R).
        \end{align}
        Combining with lemma \ref{lemmaPQR}, it holds:
        \begin{align}\label{eqvL1}
            \E_L(\E_U(\hat{v}(L_U))^2) &= \frac{1}{n^4}\left(\E(P) + 2\E(Q) + 2\E(R)\right) \nonumber \\
            &= \frac{(n-1)(n^2-2n+3)}{n^3}\mu_2^2 + \frac{(n-1)^2}{n^3}\mu_4.
        \end{align}
        Item 3. We remark again $\E_L(\E_U(\hat{v}(L_U)^2))$ is a symmetric polynomial function of $(X_i-X_j)^2$ and the degree of this polynomial is $4$. Thus there exists constants $\alpha, \beta$ and  $\gamma$ such that:
        \begin{align}\label{eqEB2}
            \E_U(\hat{v}^2(L_U)) &= \frac{1}{n^m}\frac{1}{m^2(m-1)^2}\sum_{i = 1}^{n^m} \hat{v}^2(L_{u_i}) \nonumber \\
            & = \frac{1}{n^m}\frac{1}{m^2(m-1)^2}(\alpha P+\beta Q + \gamma R).
        \end{align}
        The coefficients $\alpha, \beta$ and $\gamma$ are the coefficients of $(X_1-X_2)^4$, $(X_1-X_2)^2(X_1-X_3)^2$ and $(X_1-X_2)^2(X_3-X_4)^2$ respectively. We use similar arguments as in the proof of item 1. Denote $n_i(U)$ the number of $X_i$ in the sample of bagging $\{X_{U(j)}\}_{j=1,...,m}$. Again according to property of $G$, then it holds: 
        $$\alpha = \sum_{i = 1}^{n^m}n_1(u_i)^2n_2(u_i)^2 = n^m \partial_1^2\partial_2^2G(0,\cdots,0),$$
        $$\beta = 2\sum_{i = 1}^{n^m}n_1(u_i)^2n_2(u_i)n_3(u_i) = 2n^m \partial_1^2\partial_2\partial_3 G(0,\cdots,0),$$
        and
        $$\gamma = 2\sum_{i = 1}^{n^m}n_1(u_i)n_2(u_i)n_3(u_i)n_4(u_i) = 2n^m \partial_1\partial_2\partial_3\partial_4 G(0,\cdots,0).$$
        Therefore, with equation \eqref{fctchara} and \eqref{eqEB2}, we have
        \begin{align}\label{eqEBv2f}
        \E_U(\hat{v}^2(L_U))= &\left(\frac{1}{n^2m(m-1)} + \frac{2(m-2)}{n^3m(m-1)} + \frac{(m-2)(m-3)}{n^4m(m-1)}\right)P \nonumber\\
                      & + \left(\frac{2(m-2)}{n^3m(m-1)} + \frac{2(m-2)(m-3)}{n^4m(m-1)}\right)Q \nonumber\\
                      & + \frac{2(m-2)(m-3)}{n^4m(m-1)}R.
        \end{align}
        Item 3 follows with equation \eqref{eqEBv2f} and lemma \ref{lemmaPQR}. 
        \end{proof}}
        
        Now we are ready to prove theorem 3.1 .
        \begin{proof}[Proof of theorem 3.1 ]
        According to theorem 2.1  and lemma \ref{lemmakeyquantities}, it holds 
        \begin{equation*}
            \E_{(L,B)}(\tilde{v}) = \E_L(\E_U(\hat{v}(L_U))) = \frac{n-1}{n} \mu_2.
        \end{equation*}
        On the other hand, applying lemma \ref{lemmakeyquantities} successively, we have:
        \begin{align*}
            \E_L\big(\V_U(\hat{v}(L_U))\big) & = \E_L\left(\E_U\left(\hat{v}(L_U)^2\right)\right) - \E_L\left(\E_U(\hat{v}(L_U))^2\right) \\
            & = \frac{n-1}{nm(m-1)}\left(3m - 3 + \frac{n^2-2n+3}{n^2}(6-4m)\right)\mu_2^2 \nonumber\\
        &+ \frac{n-1}{nm(m-1)}\left(m-1+\frac{n-1}{n^2}(6-4m)\right)\mu_4,
        \end{align*}
        and:
        \begin{align*}
            \V_L\big(\E_U(\hat{v}(L_U))\big) & = \E_L\left(\E_U(\hat{v}(L_U))^2\right) - \E_L\left(\E_U(\hat{v}(L_U))\right)^2\\
            & = \frac{(n-1)(n^2-2n+3)}{n^3}\mu_2^2 + \frac{(n-1)^2}{n^3}\mu_4 - \frac{(n-1)^2}{n^2}\mu_2^2\\
            & = \frac{(3-n)(n-1)}{n^3}\mu_2^2 + \frac{(n-1)^2}{n^3}\mu_4.
        \end{align*}
        The proof is completed.
        \end{proof}

        \section{Proof of theorem 3.3} 
        
        According to theorem 2.1 and proposition 3.2, it holds
        \begin{align*}
            &\mathrm{MSE}(\tilde{v}(L,B)) - \mathrm{MSE}(\hat v)\\ 
            &= \frac{1}{N}\E_L\big(\V_U(\hat{v}(L_U))\big) + \left(\frac{5n-3}{n^3}-\frac{2}{n(n-1)}\right)\mu_2^2 + \frac{-2n+1}{n^3}\mu_4\\
            &= \frac{1}{n^2}\left(-2\mu_4+3\mu_2^2\right) + \frac{1}{n^3}\left(\mu_4-\mu_2^2\right) +  \frac{1}{N}\E_L\big(\V_U(\hat{v}(L_U))\big) + \mathcal{O}\left(\frac{1}{n^4}\right) .
        \end{align*}
        On the other hand, simplifying equation (1) from theorem 2.1, we have:
        $$\frac{1}{N}\E\big(\V_U(\hat{v}(L_U))\big) = \frac{1}{Nm}\left(\mu_4 - \mu_2^2\right) + o\left(\frac{1}{Nm}\right).$$
        We deduce therefore
        \begin{align*}
            &\mathrm{MSE}(\tilde{v}(L,B))-\mathrm{MSE}(\hat v)\nonumber \\
        &=\frac{1}{Nm}\left(\mu_4-\mu_2^2\right)+\frac{1}{n^2}\left(-2\mu_4+3\mu_2^2\right)+o\left(\frac{1}{Nm}+\frac{1}{n^2}\right).
        \end{align*}
        We deduce theorem 3.3 by comparing the latter equation with 0.

    \section{Algorithm Complexity}
    The algorithm complexity decomposes as follows: The estimation of the kurtosis takes $O(n)$, The estimation of each inner variance takes $O(n)$, thus the estimation of the bagged variance takes $O(Nn)$. The total complexity is thus $O(n) + O(Nn) \approx  O(Nn)$. Given that the condition $N > \frac{n}{2}$ must hold, we deduce that the complexity is in $O(n^2)$. Compared to traditional variance estimation in $O(n)$, the complexity is thus greater. As a result, this algorithm can only be useful in cases where the sample size $n$ is not too large. 
    
    \section{Experimental Setup Details}
    For all experiments, whenever possible we optimized and parallelized our computation using the Python Numba\footnote{\url{http://numba.pydata.org/}} library. All the experiments were run on a single CPU 3,1 GHz Intel Core i5 with 8 GB of RAM. 
    \subsection{First Experiment}
    To generate the datasets, we used the make\_regression function from sklearn\footnote{\url{https://scikit-learn.org/stable/modules/generated/sklearn.datasets.make_regression.html}}. 
    Regarding the first experiment, we used the Numpy library to estimate the parameters of the linear regression (linalg.lstsq function with default hyperparamters\footnote{\url{https://docs.scipy.org/doc/numpy-1.13.0/reference/generated/numpy.linalg.lstsq.html}}). For the Decision Tree regression, we use the scikit-learn implementation\footnote{\url{https://scikit-learn.org/stable/modules/generated/sklearn.tree.DecisionTreeRegressor.html}} with default hyperparameters. To fit the non-linear curve on the estimated data points, we used the optimize.curve\_fit function from the Scipy package\footnote{\url{https://docs.scipy.org/doc/scipy/reference/generated/scipy.optimize.curve_fit.html}}. 
    \subsection{Second Experiment}
    For the second experiment, we essentially used the random module from the Numpy library\footnote{\url{https://docs.scipy.org/doc/numpy/reference/routines.random.html}}. 
    \subsection{Third Experiment}
    For the third experiment, we used the rv\_discrete module from the Scipy library to design a custom discrete distribution\footnote{\url{https://docs.scipy.org/doc/scipy/reference/generated/scipy.stats.rv_discrete.html}}.  
    
    
    
    
    